%% file: main_preprint.tex
\documentclass[11pt, a4paper, logo, copyright]{preprint-asset/kunumi}

\usepackage[authoryear, sort&compress, round]{natbib}
\usepackage[]{mdframed}
\usepackage{anyfontsize}
\usepackage{listings}
\usepackage{hyperref}
\usepackage{url}
\usepackage{graphicx}
\usepackage{mathrsfs} %
\usepackage{etoolbox}
\usepackage{cleveref}
\usepackage{tcolorbox}
\usepackage{tabularx}
\usepackage{colortbl}
\usepackage{booktabs}       %
\usepackage{amsfonts}       %
\usepackage{nicefrac}       %
\usepackage{microtype}      %
\usepackage{subcaption}
\usepackage{multirow}
\usepackage{subcaption}
\usepackage{ragged2e}
\usepackage{enumitem}%
\setlist[itemize]{noitemsep, topsep=0pt}
\usepackage{enumitem,kantlipsum}
\usepackage{tabularx}
\usepackage{ragged2e} %
\usepackage{booktabs} %
\usepackage{xr} 
\usepackage{xcolor}

\newlength\savewidth

\definecolor{baselinecolor}{HTML}{d6eaf8}

\definecolor{mygray}{gray}{0.4}

\definecolor{darkgreen}{rgb}{0, 0.5, 0}

\AtBeginEnvironment{tcolorbox}{\tiny}

\usepackage{adjustbox}
\usepackage[utf8]{inputenc} 
\usepackage[T1]{fontenc}    
\usepackage{hyperref}       
\usepackage{url}            
\usepackage{booktabs}       
\usepackage{amsfonts}       
\usepackage{nicefrac}       
\usepackage{microtype}      
\usepackage{xcolor}         
\usepackage{xspace}
\usepackage[english]{babel}    
\usepackage{graphicx}          
\usepackage{amsmath}           
\usepackage{amssymb}           
\usepackage{multirow}          
\usepackage{rotating}          
\usepackage{soul}              
\usepackage{appendix}          
\usepackage[acronym]{glossaries}  
\usepackage{dsfont}
\usepackage{enumitem}
\usepackage{tabularx}
\usepackage{fancyhdr}

\usepackage{colortbl}
\usepackage{makecell}
\usepackage{threeparttable}
\usepackage{algorithm}
\usepackage{algpseudocode}
\usepackage[dvipsnames]{xcolor}

\newcommand{\familyrule}{
\arrayrulecolor{black}
\specialrule{0.5pt}{1pt}{1pt}
}
\newcommand{\modelrule}{
\arrayrulecolor{gray!40}
\specialrule{0.5pt}{1pt}{1pt}
}

\usepackage{float}
\input{acronyms.tex}

\title{Latent Fact-Checking: Detecting Misinformation through Activation Engineering}
\correspondingauthor{Pedro Barcelos (p.barcelos@edu.pucrs.br), Lucas S. Kupssinskü (lucas.kupssinsku@pucrs.br) and Rodrigo C. Barros (rodrigo.barros@pucrs.br) }

\author[*,1]{Pedro Barcelos}
\author[*,1]{Otávio Parraga}
\author[1]{Marcelo M. Delucis}
\author[1]{Lucas M. Fraga}
\author[1]{Lucas S. Kupssinskü} 

\author[1,2]{\\Rodrigo C. Barros}

\affil[1]{MALTA, Machine Learning Theory and Applications Lab, PUCRS, Porto Alegre, Brazil}
\affil[2]{Kunumi Institute, Brazil}

\affil[*]{\begin{minipage}{0.8\textwidth} Equal contribution\end{minipage}}

\newif\ifonecolumn
\onecolumntrue %
\begin{abstract}
\vspace{-1em}
\input{sections/abstract}
\end{abstract}

\begin{document}

\maketitle

\input{sections/1_intro}
\input{sections/2_related}
\input{sections/3_methods}
\input{sections/4_results}
\input{sections/5_conclusion}
\input{sections/z_ackowledgements}


{
\small
\bibliography{ref}
}

\end{document}

%% file: acronyms.tex
\newacronym{se}{SE}{Sentence Embeddings}
\newacronym{rag}{RAG}{Retrieval-Augmented Generation}
\newacronym{ar}{AR}{Autoregressive}
\newacronym{sts}{STS}{Semantic Textual Similarity}
\newacronym{ir}{IR}{Information Retrieval}
\newacronym{sft}{SFT}{Supervised Fine-Tuning}
\newacronym{gpu}{GPU}{Graphics Processing Unit}
\newacronym{ndcg}{NDCG}{Normalized Discounted Cumulative Gain}
\newacronym{eos}{EOS}{end-of-sequence}
\newacronym{idf}{IDF}{inverse document frequency}
\newacronym{arlm}{ARLM}{Autoregressive Language Model}
\newacronym{lm}{LM}{Language Model}
\newacronym{nlp}{NLP}{Natural Language Processing}
\newacronym{rq}{RQ}{Research Question}

\newacronym{mu}{MU}{Machine Unlearning}
\newacronym{guard}{GUARD-IT}{Inference-Time Unlearning via Gated Activation Redirection}
\newacronym{llm}{LLM}{Large Language Model}
\newacronym{sv}{SV}{Steering Vector}
\newacronym{psv}{PSV}{Prototype Steering Vector}
\newacronym{gate}{SG}{Similarity Gateway}
\newacronym{st}{ST}{Sentence-Transformer}

\newacronym{ActAdd}{ActAdd}{Activation Addition}
\newacronym{rpe}{RPE}{Representation Engineering}
\newacronym{caa}{CAA}{Contrastive Activation Addition}

\newacronym{ga}{GA}{Gradient Ascent}
\newacronym{gdiff}{GradDiff}{Gradient Difference}
\newacronym{npo}{NPO}{Negative Preference Optimization}
\newacronym{simnpo}{SimNPO}{Simple Negative Preference Optimization}
\newacronym{blur}{BLUR}{Bi-Level Unlearning}
\newacronym{rmu}{RMU}{Representation Misdirection for Unlearning}
\newacronym{cast}{CAST}{Conditional Activation Steering}
\newacronym{sadi}{SADI}{Semantics-Adaptive Dynamic Intervention}
\newacronym{undial}{UNDIAL}{Unlearning via Self-Distillation on Adjusted Logits}
\newacronym{pdu}{PDU}{Primal-Dual Unlearning}
\newacronym{satimp}{SatImp}{Saturation-Importance}
\newacronym{wga}{WGA}{Weighted Gradient Ascent}
\newacronym{ceu}{CEU}{Cross Entropy Unlearning}
\newacronym{dpo}{DPO}{Direct Preference Optimization}



%% file: sections/abstract.tex
The proliferation of misinformation online has driven demand for scalable detection systems.
While most existing approaches rely on surface-level linguistic features or external knowledge retrieval, we examine truthfulness as a geometric property of a language model's representation space.
We introduce a misinformation detection framework grounded in activation engineering, which leverages the latent geometry of transformer models. 
Our approach elicits a misinformation direction in the residual stream by contrasting activations from paired truthful and false statements, following the difference-in-means principle of Contrastive Activation Addition (CAA). 
At inference time, the last-token activation of an unseen claim is projected onto this direction, and the projected representation is fed to an Multilayer Perceptron (MLP) for classification. The procedure requires no fine-tuning of the backbone model, no external evidence retrieval, and no task-specific supervision beyond the contrastive pairs used to estimate the direction. 
We evaluate the method across $11$ models from the Gemma, Llama, and Qwen families, ranging from $270$M to $12$B parameters, on three fact-checking benchmarks: \textsc{AVeriTeC}, \textsc{LIAR}, and \textsc{FACTors}.
The falsehood direction is recoverable across model scales and architectural families, and last-token projection matches or surpasses zero-shot and few-shot prompting baselines on \textsc{LIAR} and \textsc{FACTors}, with the largest gains observed for smaller models. 
Performance on \textsc{AVeriTeC} is more limited, which we attribute to its evidence-grounded labeling scheme. 
These findings provide evidence that truthfulness is a structured, linearly separable concept in the latent space of pretrained language models, and point toward interpretability-driven misinformation detection as a practical complement to retrieval-based pipelines. 


%% file: sections/1_intro.tex
\section{Introduction}
\label{sec:intro}

Large language models (LLMs) have become the main tool of many Natural Language Processing (NLP) applications due to their fluency and contextually appropriate text across a wide range of tasks. 
That fluency, however, does not guarantee factual reliability. 
LLMs generate statements that may sound plausible while also being false or unsupported by evidence~\citep{inference_time,truthfulQA}. 
Increasing model scale alone does not solve this issue, as models still reproduce what can be described as imitative falsehoods, which are false answers that remain highly likely under the training distribution~\citep{truthfulQA}. 
In this setting, automated fact-checking (AFC) becomes necessary to detect and mitigate the spread of misinformation~\citep{guo_survey_afc}.

A limitation of standard LLM use is that veracity assessment is usually based solely on the text each model produces.
This leaves open the possibility that the model contains information relevant to factual judgment but does not express it under its default decoding behavior.
This is usually referred to as the generation-discrimination gap, where a model may possess latent knowledge that separates truth from falsehood while still failing to state the correct answer in standard generation settings~\citep{inference_time}. 
Findings from mechanistic interpretability and representation engineering support this, suggesting that models often retain internal signals related to whether a statement is true or false, and that these signals can be linearly separated in activation space~\citep{inference_time,geometry_truth,representation_engineering}. 
This motivates looking beyond final model outputs and examining internal representations directly.

Rather than relying on fine-tuning or reinforcement learning, we study factual judgment through activation engineering, which seeks to identify representations associated with high-level properties during the forward pass~\citep{steering_lms,representation_engineering}. 
Our approach combines ideas from CAA and Inference-Time Intervention (ITI).
Following Contrastive Activation Addition (CAA), we process pairs of supported and refuted claims and compute the mean activation difference as a falsehood direction~\citep{CAA}.
Claim representations are then projected onto the estimated falsehood direction at each layer, and an Multilayer Perceptron (MLP) trained on those projections is used to select the optimal layer on a holdout validation set.
A final MLP is then retrained on the full training data at the selected layer to produce the predictions~\citep{style_vectors,inference_time}.

In order to test whether our method 
is effective across architectures and scales, we evaluate it on Small Language Models (SLMs) from different families, including Llama, Qwen, and Gemma. 
Our evaluation is grounded in real fact-checking 
benchmarks, since synthetic claims often miss properties that characterize real misinformation settings~\citep{factors}. 
We therefore use \textsc{AVeriTeC}~\citep{averitec} dataset, which contains real-world claims paired with web evidence, \textsc{FACTors}~\citep{factors}, which reflects ecosystem-level fact-checking reports, and \textsc{LIAR}~\citep{liar}, which contains short political claims for veracity classification.
Within this setting, we make three contributions: $(i)$. We investigate steering vectors for factual judgment derived from fact-checking datasets; $(ii)$. We analyze transfer across datasets through comparisons of learned directions, and $(iii)$. We perform a layer-wise ablation to identify the most discriminative transformer layers for veracity judgment, and compare the resulting classifier against zero-shot and few-shot prompting baselines.

%% file: sections/2_related.tex
\section{Related Work}\label{sec:rw}
\subsection{Activation Engineering}
Activation engineering, often used interchangeably with representation engineering~\citep{representation_engineering}, is an emerging paradigm for interpreting and controlling LLM behavior without computationally intensive parameter updates.
Instead of modifying model weights, this approach directly manipulates hidden states during inference, exploiting the fact that intermediate representations encode high-level semantic concepts within a high-dimensional space.
By identifying geometric directions corresponding to these concepts, it is possible to steer model outputs at runtime through lightweight vector addition or subtraction~\citep{steering_lms}.

This approach rests on two complementary hypotheses from the mechanistic interpretability field.
The \emph{Linear Representation Hypothesis} posits that transformers encode high-level concepts as linear directions in the activation space~\citep{park2024linearrepresentationhypothesisgeometry}, this is supported empirically by the recoverable geometry of truthfulness within transformer residual streams~\citep{geometry_truth}.
Complementing this, the \emph{Superposition Hypothesis}~\citep{elhage2022superposition} proposes that neural networks represent far more features than they have dimensions by encoding distinct concepts as near-orthogonal directions within shared activation subspaces.
Together, they explain why contrastive difference-in-means methods are effective.
If concepts are linearly encoded and their directions are approximately orthogonal in superposed representations, subtracting activations elicited by opposing stimuli isolates a target concept direction without disturbing the encoding of unrelated features.

One foundational method is CAA~\citep{CAA}, which constructs paired contrastive prompts to elicit opposing behaviors, collects intermediate activations, and computes their average difference to obtain a ``steering vector''.
This vector is then scaled and injected into the model's residual stream during inference.
Beyond simple averaging, alternative approaches include training linear classifiers to define concept boundaries~\citep{interpretability}, targeting specific attention heads that correlate with a desired attribute~\citep{inference_time}, and applying dimensionality reduction over activation differences to extract dominant, noise-resistant directions~\citep{representation_engineering}.

These techniques have shown reduction in hallucinations~\citep{geometry_truth}, controlling style and sentiment~\citep{style_vectors,extracting_latent_steering,controlling_llms}, and mitigating toxic generation and adversarial jailbreaks~\citep{arditi2024refusal,personalized_steering_llms}, all at inference time, enabling rapid and context-specific adjustments.
Building on truthfulness steering, activation engineering offers an alternate direction for AFC~\citep{azaria2023internal}, where probing or steering a model's internal belief state can be integrated into systems that proactively detect and mitigate misinformation.
The veracity direction estimated here can be considered as a linearly encoded concept, recoverable from the residual stream because truthfulness occupies a consistent geometric axis across the models' superposed representations.

\subsection{Automated Fact-Checking}

AFC aims to assess the veracity of natural language claims using external evidence and reasoning models. 
Different approaches framed this task as a supervised text classification over short claims, often relying on linguistic features or metadata without explicit evidence grounding~\citep{liar}.
While effective in controlled settings, such methods are limited in their ability to generalize and provide transparent justifications. 
Recent work frames AFC as a multi-stage pipeline involving evidence retrieval and claim verification, such as \textsc{FEVER}~\citep{thorne2018fever}.
These datasets require systems to retrieve supporting evidence and reason over it, shifting the focus toward end-to-end verification. 
Subsequent benchmarks, including \textsc{FEVEROUS}~\citep{aly2019feverous}, \textsc{SciFact}~\citep{wadden2020scifact}, and \textsc{AVeriTeC}~\citep{averitec}, further increase realism by incorporating multi-hop reasoning and open-domain retrieval.

Nevertheless, current AFC approaches still face challenges~\citep{stil_paper_afc}.
Prompting-based strategies require large models and are unstable across phrasings; fine-tuned classifiers demand substantial labeled data and generalize poorly across domains.
Evidence-grounded systems rely on costly retrieval infrastructure sensitive to evidence quality.
Our method needs no fine-tuning, no external retrieval, and no large-scale labeled corpora.
By operating directly on the internal activations of small frozen language models, it offers a lightweight, generalizable alternative for claim-level veracity assessment.

\subsection{LLM-based Misinformation Detection}

LLMs have been increasingly applied to misinformation and fake news detection through prompting strategies~\citep{guo_survey_afc} and fine-tuning~\citep{zeng2024factchecking,pavlyshenko2023analysis}. 
Zero-shot and few-shot prompting offer deployment flexibility but exhibit limitations.
Models frequently hallucinate plausible-sounding justifications~\citep{guo_survey_afc}, and outputs are highly sensitive to prompt wording, with minor rephrasing causing large swings in accuracy on identical claims.
State-of-the-art systems on the \textsc{AVeriTeC} shared task~\citep{averitec} demonstrate that retrieval-augmented generation and zero-shot prompting can achieve competitive performance without fine-tuning~\citep{ullrich2024aic,mohammadkhani2024zeroshot}, yet these approaches rely on external access to information.
Fine-tuning mitigates some surface-level failures but demands large labeled datasets and generalizes poorly across domains and claim types.
Together, these limitations suggest that surface-level generation is an unreliable proxy for veracity assessment.
Rather than querying model outputs, our approach bypasses generation entirely and reads the internal representations where veracity signals are latently encoded, motivating the activation-based classification framework developed in this work.

%% file: sections/3_methods.tex
\section{Methodology}

\subsection{Problem Formulation}

Let $\mathcal{D} = \{(s_i, y_i)\}_{i=1}^{N}$ be a misinformation dataset, where $s_i$ is a natural language claim and $y_i \in \{0, 1\}$ is its falsehood label ($y_i = 0$ for true, $y_i = 1$ for false). 
We consider a transformer-based language model $\mathcal{M}$ with $L$ layers whose residual stream at layer $\ell$ produces a hidden representation $\mathbf{h}^{(\ell)}_t \in \mathbb{R}^d$ for each token position $t$, where $d$ is the model's hidden dimension. 
Our goal is to learn a classifier $f: \mathbb{R}^d \rightarrow \{0,1\}$ that operates directly on these internal representations, without any gradient-based fine-tuning of $\mathcal{M}$.

\subsection{Contrastive Prompt Construction}

For each training instance $s_i$, we construct a pair of structured prompts 
that anchor the completion to a specific veracity polarity through a forced \texttt{A/B} choice:

\begin{align}
    p^{+}_i &= \texttt{``}s_i\texttt{.\textbackslash nThe previous assertion is (A) True (B) False\textbackslash nAnswer: (A)''} \\
    p^{-}_i &= \texttt{``}s_i\texttt{.\textbackslash nThe previous assertion is (A) True (B) False\textbackslash nAnswer: (B)''}
\end{align}

\noindent where $p^{+}_i$ anchors the answer to truth and $p^{-}_i$ anchors it to falsehood. 

Both prompts are derived from the same $s_i$, so the only variable between 
them is the polarity of the forced answer token. This controls for topic, 
syntax, and length. This prompt addition is only used for the misinformation direction estimation (Section~\ref{sec:dir}). 

To prevent the model from associating misinformation with a 
fixed letter, we randomize the mapping between \texttt{A}/\texttt{B} and 
True/False across instances, with each assignment drawn uniformly and independently for every $s_i$.

\subsection{Activation Extraction}

We pass each prompt through the frozen model $\mathcal{M}$ and extract the last-token hidden state at layer $\ell$.
In a causally masked transformer, the last token aggregates contextual information from the entire preceding sequence and therefore serves as a natural sentence-level representation. 
Formally:

\begin{equation}
    \mathbf{h}^{(\ell)\pm}_i = \texttt{LastTok}\!\left(\mathcal{M}^{(\ell)}(p^{\pm}_i)\right) \in \mathbb{R}^d
\end{equation}

All layer activations are collected in a single forward pass via hooks, incurring no additional inference overhead.
In addition to the contrastive-prompt activations $\mathbf{h}^{(\ell)\pm}_i$, we also extract last-token activations from the bare claims $s_i$—without any annotation template—at every layer; these plain-claim activations are used for projection and classification.

\subsection{Falsehood Direction Estimation}\label{sec:dir}

We use contrastive difference-in-means \citep{CAA} to identify a misinformation direction in the residual stream, and search for the specific layers where the target attribute can be better steered.
The falsehood direction at layer $\ell$ is defined as the mean difference between the truthful and false activation distributions from the prompts of the training datasets with refuted samples:

\begin{equation}
    \mathbf{v}^{(\ell)} = \frac{1}{N}\sum_{i=1}^{N} \mathbf{h}^{(\ell)+}_i \;-\; \frac{1}{N}\sum_{i=1}^{N} \mathbf{h}^{(\ell)-}_i
\end{equation}

Under the linear representation hypothesis~\citep{park2024linearrepresentationhypothesisgeometry}, truthfulness and falsehood correspond to separable directions in the residual stream; $\mathbf{v}^{(\ell)}$ is our estimate of that axis.
While CAA uses this direction as a steering vector to be added to the residual stream during generation, we instead retain it as a fixed reference axis: bare claim activations are projected onto $\hat{v}^{(\ell)}$ for binary classification, leaving the model's forward pass unmodified.
We normalize it as $\hat{\mathbf{v}}^{(\ell)} = \mathbf{v}^{(\ell)} / \|\mathbf{v}^{(\ell)}\|_2$ for use in downstream projection.

\subsection{Vector Projection and MLP Training}\label{sec:proj}

Given the normalized misinformation direction $\hat{\mathbf{v}}^{(\ell)}$ estimated at layer $\ell$ (Section~\ref{sec:dir}), we project each training claim's last-token activation—extracted from the bare claim $s_i$ without any \texttt{A/B} template—onto this direction:

\begin{equation}
    \mathbf{z}^{(\ell)}_i = \left(\mathbf{h}^{(\ell)}_i \cdot \hat{\mathbf{v}}^{(\ell)}\right)\hat{\mathbf{v}}^{(\ell)} \in \mathbb{R}^d, \quad \mathbf{h}^{(\ell)}_i = \texttt{LastTok}\!\left(\mathcal{M}^{(\ell)}(s_i)\right)
\end{equation}

A large magnitude of $\mathbf{z}^{(\ell)}_i$ indicates strong alignment with the misinformation direction.
Using untemplated claims ensures the classifier generalizes to unannotated inputs at deployment.

We then train a simple MLP classifier with one hidden layer with a dimension of 256, Adam optimizer, and a learning rate of 0.001 for 200 epochs on the projected training vectors:
\begin{equation}
    \hat{y}_i = \text{MLP}^{(\ell)}\!\left(\mathbf{z}^{(\ell)}_i\right)
\end{equation}

This projection–MLP pair is instantiated independently for every layer $\ell$, enabling the layer selection procedure described next.

\subsection{Layer Selection}\label{sec:layer}

Since the linear separability of falsehood representations varies across layers, we select the optimal layer $\ell^*$ using a held-out validation split $\mathcal{D}_\text{val}$.
For each layer $\ell$, we apply the procedure of Section~\ref{sec:proj}: project the plain-claim activations of $\mathcal{D}_\text{train}$ onto $\hat{\mathbf{v}}^{(\ell)}$, train an MLP on the resulting vectors, and evaluate it on the plain-claim activations of $\mathcal{D}_\text{val}$ projected onto the same direction.
The layer that maximizes validation accuracy is retained:

\begin{equation}
    \ell^* = \underset{\ell \in \{1, \ldots, L\}}{\arg\max} \;\; \text{Acc}\!\left(\text{MLP}^{(\ell)},\; \mathcal{D}_\text{val}\right)
\end{equation}

\subsection{Final Model}

Once $\ell^*$ is identified, we pool all available labeled data $\mathcal{D}_\text{train} \cup \mathcal{D}_\text{val}$ to obtain a more representative misinformation direction. We recompute $\hat{\mathbf{v}}^{(\ell^*)}$ following Section~\ref{sec:dir} using contrastive prompts built from the combined split, project every plain-claim activation at layer $\ell^*$ onto this updated direction, and retrain the MLP from scratch on the resulting vectors. The MLP weights produced during layer selection are discarded; no data leakage occurs since those weights play no role in the final classifier.

To classify a new unseen claim $s$ at inference time, we extract its last-token activation at layer $\ell^*$ without any template, project it onto $\hat{\mathbf{v}}^{(\ell^*)}$, and feed the result to the final MLP:

\begin{equation}
    \hat{y} = \text{MLP}\!\left(\left(\mathbf{h}^{(\ell^*)} \cdot \hat{\mathbf{v}}^{(\ell^*)}\right)\hat{\mathbf{v}}^{(\ell^*)}\right), \quad \mathbf{h}^{(\ell^*)} = \texttt{LastTok}\!\left(\mathcal{M}^{(\ell^*)}(s)\right)
\end{equation}

This requires only a single forward pass through the frozen model and no annotation template, making the approach directly applicable to unannotated claims at deployment.

\subsection{Datasets}

We consider three complementary benchmarks widely used in AFC: \textsc{AVeriTeC} \citep{averitec}, \textsc{LIAR}~\citep{liar}, and \textsc{FACTors}~\citep{factors}. 
Together, these datasets represent distinct axes of the problem scope, including open-domain verification, domain-specific classification, and large-scale ecosystem analysis.

\paragraph{\textsc{AVeriTeC} (Automated Verification of Real-world Claims with Evidence from the Web)} is a dataset designed to reflect realistic fact-checking pipelines~\citep{averitec}. 
It contains $4,568$ real-world claims collected from over $50$ fact-checking organizations via sources such as the Google Fact Check API. 
A key characteristic is its open-domain setting, which covers diverse topics including politics, health, science, and current events. 
Unlike other datasets, it explicitly models the multi-step nature of verification by providing question–answer (QA) pairs, retrieved web evidence, and textual justifications. 
This structure requires systems to jointly perform claim decomposition, external evidence retrieval, and reasoning, making it a challenging benchmark for end-to-end AFC.

\paragraph{\textsc{LIAR}}~\citep{liar} is a consolidated dataset constructed from claims annotated by the \texttt{PolitiFact} fact-checking platform.
It consists of approximately $12,800$ short statements sourced from speeches, debates, and campaign materials, each labeled using a fine-grained truthfulness scale (e.g., \textit{True}, \textit{Half True}, \textit{False}, \textit{Pants on Fire}). 
In contrast to open-domain benchmarks, \textsc{LIAR} categorizes fact-checking primarily as a supervised text classification task, without requiring explicit evidence retrieval. 
While its constrained political domain and short claim format simplify modeling, prior work has identified artifacts such as over-reliance on speaker metadata and limited contextual grounding. 
Despite these limitations, it remains a standard dataset for evaluating classification-based approaches.

\paragraph{\textsc{FACTors}} is a recently introduced large-scale dataset that captures the structure of the fact-checking ecosystem~\citep{factors}.
It comprises over $118,000$ claims extracted from approximately $117,000$ fact-checking reports published by $39$ organizations over multiple decades.
Similar to \textsc{AVeriTeC}, it is multi-domain, covering topics such as politics, health, science, and general misinformation.
However, its distinguishing feature lies in the inclusion of metadata describing fact-checking organizations, authorship, and inter-claim relationships. 
This enables analyses beyond individual claim verification, such as modeling institutional behavior, credibility patterns, and cross-source agreement. 
As a result, \textsc{FACTors} supports research on large-scale, context-aware, and socially grounded fact-checking systems.

\paragraph{Experimental Setup}

For each dataset, claims are sampled from the original data and partitioned into two splits: $200$ instances are used for training and hyperparameter tuning, including layer selection, and $100$ instances are held out for evaluation. Of the $200$ training instances, $70\%$ ($140$) are used for MLP training and direction estimation, while the remaining $30\%$ ($60$) form the validation split for layer selection. 
To enable a unified binary classification framework, all label spaces are coarse-grained into two classes.
For \textsc{LIAR}, the original six-level truthfulness scale is collapsed into \textit{True} (comprising \textit{True}, \textit{Mostly True}, and \textit{Half True}) and \textit{False} (comprising \textit{Barely True}, \textit{False}, and \textit{Pants on Fire}). 
For \textsc{AVeriTeC} and \textsc{FACTors}, the original binary labels are retained without any modification.


%% file: sections/4_results.tex
\section{Results and Discussion} 
\label{sec:results}

In Table~\ref{tab:method_comparison}, we report the accuracy and F1 scores for our method alongside zero- and few-shot prompting baselines that we implemented and ran. 
We conducted these evaluations across eleven models—spanning the Gemma-3, Llama-3, and Qwen3.5 architectural families—and three datasets of distinct nature: \textsc{AVeriTeC}~\citep{averitec}, \textsc{LIAR}~\citep{liar}, and \textsc{FACTors}~\citep{factors}.

The models range from $270M$ to $12B$ parameters, which allows us to assess whether the falsehood direction in the latent space is recoverable across scales.
Since our work targets LLM-driven misinformation detection through a lightweight pipeline that requires no training or gradient updates, zero-shot and few-shot prompting serve as the natural reference class for methods that rely solely on model outputs, making them appropriate baselines against which to evaluate our approach.

Our results reveal a striking and consistent pattern: our latent falsehood direction outperforms both prompt-driven baselines on \textsc{LIAR} and \textsc{FACTors} across virtually all model configurations and scales.
Both datasets are relatively balanced in their binary label distribution after coarse-graining, which facilitates a well-centered difference-in-means estimate and supports the coherent falsehoods we observe across model families.
On \textsc{LIAR}, the gains are most pronounced for smaller models, where Gemma-3-4B improves from $0.57$ to $0.79$ accuracy, Llama-3.2-1B from $0.46$ to $0.73$, and Gemma-3-1B from $0.52$ to $0.75$, which represents relative gains of over $30\%$.
Similar gains are observed on \textsc{FACTors}: Llama-3.2-1B improves from $0.47$ to $0.70$, Qwen3.5-2B from $0.54$ to $0.68$, and Qwen3.5-9B from $0.64$ to $0.71$.
These improvements extend to the smallest models evaluated: even Qwen3.5-0.8B, which performs near chance under zero-shot prompting on \textsc{LIAR} ($0.48$ Acc), reaches $0.68$ Acc with our method.
Moreover, the latent method consistently outperforms few-shot prompting despite few-shot providing explicit label exemplars, a finding that underscores the core contribution of our work.

Our results support the hypothesis that LLMs' internal representations encode a veracity signal that is richer and more discriminative than what is expressed through surface-level text generation.
Even when prompting fails to elicit the correct judgment, particularly at small scales, the latent falsehood recovers a discriminative structure from the residual stream, suggesting that models may already encode information relevant to factual judgment without surfacing it in their outputs~\citep{azaria2023internal}.

\input{tables/results_table}

\textsc{AVeriTeC} constitutes the main exception: here, our method performs comparably or slightly below the baselines for larger models, though it remains competitive or superior for smaller ones (e.g., Llama-3.2-1B: $0.44$ to $0.61$ Acc; Qwen3.5-4B: $0.60$ to $0.66$ Acc).
Beyond its evidence-dependent labeling scheme, \textsc{AVeriTeC} exhibits a higher proportion of refuted claims and greater variance in claim length and topic scope than \textsc{LIAR} or \textsc{FACTors}, which may bias the mean activations and partially explain the noisier falsehood direction observed in this setting.
We attribute the broader underperformance to a structural property of \textsc{AVeriTeC}: its veracity labels are determined not by the content of the claim itself, but by multi-step chains of external evidence retrieved from the web.
A claim may be entirely plausible in isolation, yet labeled \textit{Refuted} based on facts that only become accessible through document retrieval and multi-hop reasoning~\citep{averitec}.

Since our method operates solely on the last-token activation of the unaided claim text, without access to retrieved evidence or justification chains, it cannot recover the evidence-grounded judgment that \textsc{AVeriTeC} demands.
Larger prompting baselines can partially compensate through their richer parametric knowledge, substituting learned factual associations for explicit retrieval, hence their relative advantage at larger scales.
This structural mismatch between a claim-only representation and an evidence-dependent label is therefore an inherent limitation of any activation-only approach applied to retrieval-grounded benchmarks, and points toward the integration of retrieval-augmented representations as a natural direction for future work.
Taken together, our findings suggest that truthfulness is encoded as a linearly separable direction in the residual stream of transformer models across families and scales, and that this structure can be reliably exploited for misinformation detection without fine-tuning, as long as the veracity signal is grounded in the claim text itself.

\subsection{Out-of-distribution Cases}

\begin{figure}[t]
  \centering
  \makebox[\textwidth][c]{\includegraphics[width=1.0\textwidth]{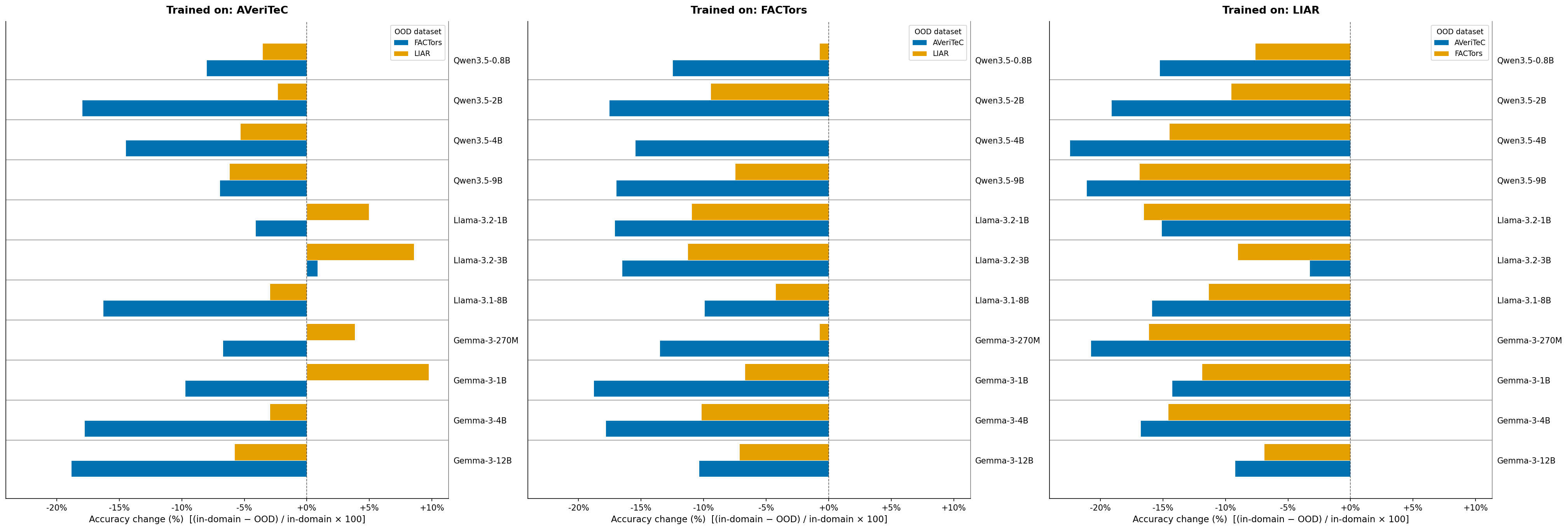}}
  \caption{Relative accuracy drop (\%) when the steering vector trained on one dataset is applied out-of-distribution (OOD) to the other two datasets, across eleven models spanning three architectural families. Bars pointing left indicate degradation; bars at or to the right of zero indicate neutral or positive transfer.}
  \label{fig:oop_drop}
\end{figure}

A natural question is whether the falsehood direction extracted from one dataset generalizes to claims from a different distribution.
Figure~\ref{fig:oop_drop} summarizes the accuracy change when the misinformation direction
vector computed on one dataset is applied to held-out claims from the remaining two datasets, expressed as the relative drop in accuracy with respect to the in-domain baseline.

Our results reveal a clear asymmetry across training conditions.
When the vector is trained on \textsc{AVeriTeC}, OOD performance degrades: drops of up to $20\%$ are observed on \textsc{FACTors} for models such as Gemma-3-12B and Llama-3.1-8B, and transfers to \textsc{LIAR} are inconsistent, with some models exhibiting near-zero change and others showing modest positive transfer.
This instability is consistent with \textsc{AVeriTeC} being an evidence-dependent dataset, whose claims are tightly coupled to retrieved documents, making the extracted direction less portable.
In contrast, vectors trained on \textsc{FACTors} or \textsc{LIAR} generalize symmetrically: cross-transfer drops between these two datasets are moderate ($5$--$10\%$) and relatively consistent across model families, suggesting that statement-only, label-balanced datasets yield more transferable falsehood directions.

Across all conditions, larger models tend to exhibit smaller relative degradation, indicating that richer internal representations produce more robust cross-dataset generalizations~\citep{CAA,representation_engineering}.
Our findings point to dataset construction choices, particularly evidence dependence and label balance, as key determinants of the portability of CAA-based falsehood directions.

\subsection{Instance-Accuracy Relation}

We also investigate the impact of the sample size used during the extraction phase on the classifier's overall performance.
Specifically, we vary the number of paired instances, denoted as $N$, utilized to compute the falsehood direction vector and to train the MLP classifier.

\begin{figure}[t]
  \centering
  \makebox[\textwidth][c]{\includegraphics[width=1.0\textwidth]{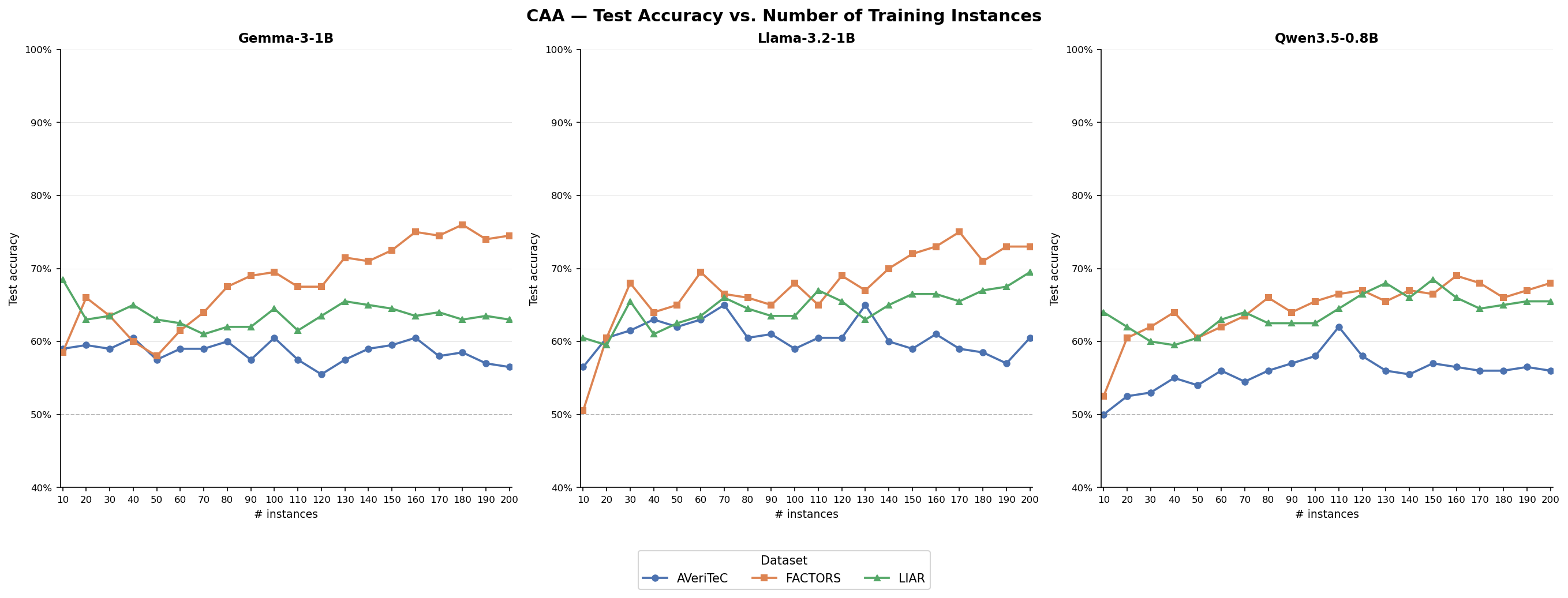}}
  \caption{Test accuracy as a function of the number of contrastive training instances $N$ used to compute the falsehood direction vector, evaluated on the three models with a similar number of parameters across three datasets. \textsc{FACTors} and \textsc{LIAR} show a positive scaling trend, while \textsc{AVeriTeC} remains flat.}
  \label{fig:instance_accuracy}
\end{figure}

Figure~\ref{fig:instance_accuracy} illustrates this trend across datasets for the three models evaluated that have around 1B parameters (Gemma-3-1B, Llama-3.2-1B, and Qwen3.5-0.8B).
The results reveal a clear dataset-dependent scaling behavior.
For \textsc{FACTors}, accuracy rises from approximately $60\%$ at $N{=}10$ to around $75\%$ at $N{=}200$, a gain of roughly $10$--$20$ percentage points across all three architectures.
\textsc{LIAR} exhibits a shallower upward trend, improving from $\sim$$60$--$65\%$ to $\sim$$65$--$70\%$ over the same range.
\textsc{AVeriTeC}, by contrast, remains essentially flat throughout ($55$--$60\%$), showing no meaningful benefit from additional contrastive pairs.
This three-way pattern is reproduced across all model families shown, indicating that the scaling behavior is driven by dataset structure rather than model-specific factors.

We attribute \textsc{AVeriTeC}'s stagnation to its structural heterogeneity: claims span highly diverse topics and require evidence-grounded reasoning, so increasing the number of contrastive pairs does not converge to a single stable falsehood direction; different claims pull the mean activation toward different semantic regions.
For claim-homogeneous datasets such as \textsc{FACTors} and \textsc{LIAR}, the difference-in-means estimator benefits substantially from larger $N$, as additional pairs more effectively cancel noise and context-specific artifacts, yielding a more robust and generalizable representation.
Addressing \textsc{AVeriTeC}'s heterogeneity may require topic-conditioned steering directions rather than a single global vector.

%% file: tables/results_table.tex
\begin{table*}[t!]
\caption{Performance comparison across datasets (Accuracy / F1).}
\centering
\small
\setlength{\tabcolsep}{5pt}
\renewcommand{\arraystretch}{1.2}

\begin{tabular*}{\textwidth}{@{\extracolsep{\fill}}llcccccc}
\toprule
\multirow{2}{*}{\textbf{Model}} 
& \multirow{2}{*}{\textbf{Method}} 
& \multicolumn{2}{c}{AVeriTeC} 
& \multicolumn{2}{c}{FACTors} 
& \multicolumn{2}{c}{LIAR} \\
\cmidrule(lr){3-4} \cmidrule(lr){5-6} \cmidrule(lr){7-8}

& 
& Acc & F1 
& Acc & F1 
& Acc & F1 \\

\midrule

\multirow{3}{*}{Gemma3-12B}
& Zero-shot & \textbf{0.72} & \textbf{0.71} & 0.68 & 0.68 & 0.64 & \textbf{0.64} \\
& Few-shot  & 0.71 & \textbf{0.71} & 0.69 & 0.69 & 0.64 & \textbf{0.64} \\
& Our Method& 0.69 & 0.67 & \textbf{0.77} & \textbf{0.77} & \textbf{0.65} & \textbf{0.64} \\

\modelrule

\multirow{3}{*}{Gemma3-4B}
& Zero-shot & \textbf{0.70} & \textbf{0.69} & 0.57 & 0.57 & 0.63 & 0.63 \\
& Few-shot  & 0.69 & 0.68 & 0.60 & 0.57 & 0.64 & 0.62 \\
& Our Method& 0.68 & 0.68 & \textbf{0.79} & \textbf{0.79} & \textbf{0.69} & \textbf{0.66} \\

\modelrule

\multirow{3}{*}{Gemma3-1B}
& Zero-shot & 0.56 & \textbf{0.54} & 0.52 & 0.51 & 0.54 & 0.47 \\
& Few-shot  & 0.51 & 0.36 & 0.50 & 0.35 & 0.52 & 0.38 \\
& Our Method& \textbf{0.57} & \textbf{0.54} & \textbf{0.75} & \textbf{0.75} & \textbf{0.63} & \textbf{0.63} \\

\modelrule

\multirow{3}{*}{Gemma3-270M}
& Zero-shot & \textbf{0.54} & \textbf{0.53} & 0.54 & 0.54 & 0.48 & 0.48 \\
& Few-shot  & \textbf{0.54} & 0.48 & 0.44 & 0.38 & 0.48 & 0.36 \\
& Our Method& 0.52 & 0.51 & \textbf{0.67} & \textbf{0.67} & \textbf{0.65} & \textbf{0.65} \\

\familyrule

\multirow{3}{*}{LLaMA3.1-8B}
& Zero-shot & 0.62 & 0.62 & 0.62 & 0.62 & 0.58 & 0.56 \\
& Few-shot  & \textbf{0.69} & \textbf{0.68} & 0.67 & 0.66 & 0.65 & 0.64 \\
& Our Method& 0.68 & \textbf{0.68} & \textbf{0.71} & \textbf{0.73} & \textbf{0.66} & \textbf{0.67} \\

\modelrule

\multirow{3}{*}{LLaMA3.2-3B}
& Zero-shot & 0.46 & 0.42 & 0.52 & 0.47 & 0.51 & 0.45 \\
& Few-shot  & \textbf{0.64} & \textbf{0.63} & 0.62 & 0.62 & \textbf{0.64} & \textbf{0.62} \\
& Our Method& 0.59 & 0.56 & \textbf{0.67} & \textbf{0.69} & 0.61 & 0.59 \\
\modelrule

\multirow{3}{*}{LLaMA3.2-1B}
& Zero-shot & 0.44 & 0.40 & 0.46 & 0.40 & 0.47 & 0.40 \\
& Few-shot  & 0.50 & 0.50 & 0.61 & 0.61 & 0.46 & 0.45 \\
& Our Method& \textbf{0.61} & \textbf{0.56} & \textbf{0.73} & \textbf{0.74} & \textbf{0.70} & \textbf{0.70} \\

\familyrule

\multirow{3}{*}{Qwen3.5-9B}
& Zero-shot & 0.68 & 0.68 & 0.56 & 0.56 & 0.64 & 0.62 \\
& Few-shot  & \textbf{0.71} & \textbf{0.71} & 0.57 & 0.56 & 0.62 & 0.60 \\
& Our Method& 0.65 & 0.63 & \textbf{0.74} & \textbf{0.75} & \textbf{0.71} & \textbf{0.71} \\

\modelrule

\multirow{3}{*}{Qwen3.5-4B}
& Zero-shot & 0.60 & 0.54 & 0.53 & 0.47 & 0.56 & 0.47 \\
& Few-shot  & 0.60 & 0.56 & 0.55 & 0.51 & 0.57 & 0.49 \\
& Our Method& \textbf{0.66} & \textbf{0.62} & \textbf{0.71} & \textbf{0.72} & \textbf{0.69} & \textbf{0.68} \\

\modelrule

\multirow{3}{*}{Qwen3.5-2B}
& Zero-shot & 0.60 & 0.57 & 0.48 & 0.42 & 0.54 & 0.47 \\
& Few-shot  & 0.58 & 0.56 & 0.58 & 0.57 & 0.59 & 0.55 \\
& Our Method& \textbf{0.64} & \textbf{0.60} & \textbf{0.74} & \textbf{0.75} & \textbf{0.68} & \textbf{0.67} \\

\modelrule

\multirow{3}{*}{Qwen3.5-0.8B}
& Zero-shot & 0.54 & 0.50 & 0.48 & 0.44 & 0.56 & 0.54 \\
& Few-shot  & 0.51 & 0.44 & 0.50 & 0.43 & 0.50 & 0.40 \\
& Our Method& \textbf{0.56} & \textbf{0.54} & \textbf{0.68} & \textbf{0.70} & \textbf{0.66} & \textbf{0.66} \\

\familyrule
\end{tabular*}
\label{tab:method_comparison}
\end{table*}

%% file: sections/5_conclusion.tex
\section{Conclusion and Future Work}\label{sec:conclusion}

Our work presents a lightweight, fine-tuning-free approach to misinformation detection grounded in activation engineering and all code is available at \url{https://github.com/Malta-Lab/LaFaCt}.
By extracting a contrastive falsehood direction from the residual stream of frozen transformer models and projecting claim representations onto it, our method consistently outperforms zero-shot and few-shot prompting baselines on \textsc{LIAR} and \textsc{FACTors} across $11$ models ranging from $270M$ to $12B$ parameters.
Gains are especially pronounced for smaller models, demonstrating that the latent falsehood signal can compensate for limited model capacity, which makes our approach particularly practical for settings where large-scale inference is not feasible.

A key finding is that the method's internal view of a claim often captures falsehood better than what a model produces through generation, consistent with the hypothesis that LLMs encode truth-related information that does not fully surface in their outputs~\citep{azaria2023internal}.
These results position our approach as a complementary layer in AFC pipelines: it does not replace retrieval-based verification or evidence-grounded reasoning, but provides an additional, computationally inexpensive signal that can be combined with existing systems.

The main limitation identified is the gap in performance on \textsc{AVeriTeC}, where falsehood depends on external evidence rather than claim text alone.
This structural mismatch motivates two concrete directions for future work: $(i)$. integrating retrieved document representations into the activation extraction step, so the falsehood direction is conditioned on evidence rather than the bare claim; and $(ii)$. developing topic-aware or domain-gated mechanisms that estimate separate falsehood directions per cluster of claims, addressing the heterogeneity that degrades the difference-in-means estimator under distributional diversity.

Additionally, our cross-dataset transfer experiments reveal that the generalizability of the falsehood direction improves with model scale, suggesting that further investigation of layer-wise representation quality across model sizes could identify more universal veracity subspaces.
Collectively, these extensions would move the approach from a claim-only classifier toward a full fact-checking component capable of operating over heterogeneous, evidence-dependent claim distributions.

%% file: sections/z_ackowledgements.tex
\subsubsection*{Acknowledgments}
\label{chap:ack}
This study was financed in part by the Coordination for the Improvement of Higher Education Personnel (CAPES) — Finance Code 001; 
by Conselho Nacional de Desenvolvimento Científico e Tecnológico (CNPq) — Grant Number: 443072/2024-8; 
and by Fundação de Amparo à Pesquisa do Estado do Rio Grande do Sul (FAPERGS) — Grant Number: 25/2551-0000891-3 and 25/2551-0002824-8.
This paper was achieved in cooperation with HP Brasil Indústria e Comércio de Equipamentos Eletrônicos LTDA. using incentives of Brazilian Informatics Law (Law nº 8.2.48 of 1991).
This work was supported by Kunumi Institute. 
The authors thank the institution for its financial support and commitment to advancing scientific research.